\documentclass[letterpaper, 10 pt, conference]{ieeeconf}

\IEEEoverridecommandlockouts 
\usepackage{graphicx} 
\usepackage{amsmath} 
\usepackage{amsfonts} 
\usepackage{bbm} 
\usepackage[table,xcdraw]{xcolor} 
\usepackage{pifont} 
\usepackage{multirow} 
\usepackage[noadjust]{cite}

\definecolor{citecolor}{HTML}{0072BC}  
\definecolor{linkcolor}{HTML}{0072BC}  
\definecolor{urlcolor}{HTML}{0072BC}  
\definecolor{customgreen}{HTML}{228B22}  
\definecolor{customred}{HTML}{FF2E17}  

\renewcommand{\baselinestretch}{1} 
\def\thickhline{\noalign{\hrule height.9pt}}

\renewcommand{\vec}[1]{\boldsymbol{#1}}
\newcommand{\mat}[1]{\boldsymbol{#1}}

\newcommand{\set}[1]{\mathcal{#1}}
\newcommand{\R}{\mathbb{R}}

\title{\LARGE \bf Depth3DLane: Fusing Monocular 3D Lane Detection with Self-Supervised Monocular Depth Estimation}

\author{
    Max van den Hoven$^{1, *}$, Kishaan Jeeveswaran$^{1}$, Pieter Piscaer$^{2}$, \\Thijs Wensveen$^{2}$, Elahe Arani$^{1}$, Bahram Zonooz$^{1}$
    \thanks{$^{1}$Max van den Hoven, Kishaan Jeeveswaran, Elahe Arani, and Bahram Zonooz are with  Eindhoven University of Technology, 5612 AZ Eindhoven, The Netherlands.}%
    \thanks{$^{2}$Pieter Piscaer and Thijs Wensveen are with TNO, 2597 AK The Hague, The Netherlands.}%
    \thanks{$^{*}$ Corresponding author: {\tt\small max.hoven@gmail.com}}%
}

\begin{document}

\maketitle
\thispagestyle{empty}
\pagestyle{empty}

\begin{abstract}
Monocular 3D lane detection is essential for autonomous driving, but challenging due to the inherent lack of explicit spatial information. Multi-modal approaches rely on expensive depth sensors, while methods incorporating fully-supervised depth networks rely on ground-truth depth data that is impractical to collect at scale. Additionally, existing methods assume that camera parameters are available, limiting their applicability in scenarios like crowdsourced high-definition (HD) lane mapping. To address these limitations, we propose Depth3DLane, a novel dual-pathway framework that integrates self-supervised monocular depth estimation to provide explicit structural information, without the need for expensive sensors or additional ground-truth depth data. Leveraging a self-supervised depth network to obtain a point cloud representation of the scene, our bird's-eye view pathway extracts explicit spatial information, while our front view pathway simultaneously extracts rich semantic information. Depth3DLane then uses 3D lane anchors to sample features from both pathways and infer accurate 3D lane geometry. Furthermore, we extend the framework to predict camera parameters on a per-frame basis and introduce a theoretically motivated fitting procedure to enhance stability on a per-segment basis. Extensive experiments demonstrate that Depth3DLane achieves competitive performance on the OpenLane benchmark dataset. Furthermore, experimental results show that using learned parameters instead of ground-truth parameters allows Depth3DLane to be applied in scenarios where camera calibration is infeasible, unlike previous methods.
\end{abstract}

\section{INTRODUCTION}
The goal of monocular 3D lane detection is to utilize a single forward-facing camera to identify lane lines and road boundaries around the ego-vehicle. These detections form the basis for various downstream tasks, such as lane departure warning~\cite{Narote2018}, high-definition (HD) mapping~\cite{Liu2020}, and automated infrastructure inspection~\cite{AIDrivenMaintenance2021, AIDrivenMaintenanceV22022}. While many approaches have been proposed for monocular 3D lane detection, inferring 3D structure remains a fundamental challenge due to the inherent lack of explicit spatial information~\cite{BEV-LaneDet2023}.

A common approach is to integrate inverse perspective mapping (IPM)~\cite{IPM1991} into the network architecture to project features to the bird's-eye view (BEV) under the flat ground assumption~\cite{3D-LaneNet2019, ReconstructFromBEV2022, BEV-LaneDet2023}. However, this assumption leads to non-negligible distortions when broken, as is often the case in uphill and downhill driving scenarios~\cite{Anchor3DLane2023, LATR2023}. BEV-based methods leverage these distortions by inferring the height of lanes above the ground plane from the vector field that moves distorted lanes to their true (parallel) location in the BEV~\cite{Gen-LaneNet2020}. However, BEV-based methods typically fail to learn robust representations, as important structural cues above the road surface are severely distorted during IPM~\cite{Anchor3DLane2023}. To address this issue, recent methods propose to avoid IPM entirely and rely on projecting 3D lane proxies to the front view to sample relevant features~\cite{Anchor3DLane2023, LATR2023}. Despite advancing state-of-the-art, these BEV-free methods lack explicit spatial information to infer 3D lane geometry from~\cite{M2-3DLaneNet2022}.

Multi-modal methods, such as M$^2$-3DLaneNet~\cite{M2-3DLaneNet2022}, rely on specialized depth sensors to produce explicit spatial information, but these are prohibitively expensive for many applications~\cite{Lite-Mono2023}. Other methods, such as SALAD~\cite{ONCE-3DLanes2022}, incorporate fully-supervised depth networks to predict 3D structure, but these rely on ground-truth depth data that is impractical to collect at scale~\cite{Zhou2017, Godard2019}. Furthermore, current methods assume that accurate camera parameters are available~\cite{Ma2024}, limiting applications where camera calibration is infeasible, such as crowdsourced HD lane mapping~\cite{AIDrivenMaintenance2021}.

To address the aforementioned limitations of previous methods, we propose Depth3DLane, a novel dual-pathway framework that integrates self-supervised monocular depth estimation to predict explicit spatial information, without relying on expensive sensors or additional ground-truth depth data. Self-supervised monocular depth estimation methods learn to jointly predict dense depth and ego-motion, such that the 3D structure of a scene is explained consistently across a sequence of unlabeled monocular video frames. By combining explicit spatial information from the bird's-eye view pathway with rich semantic information from the front view pathway, Depth3DLane is able to accurately infer 3D lane geometry.

Furthermore, we show that our framework can be extended to learn camera intrinsics in a self-supervised manner. To address the instability of per-frame learned intrinsics, we propose a theoretically motivated fitting procedure to estimate intrinsics on a per-segment basis. This approach allows our framework to be applied in scenarios where camera calibration is infeasible, unlike previous methods.

Our contributions can be summarized as follows:
\begin{enumerate}
    \item We propose Depth3DLane, a dual-pathway framework that integrates monocular 3D lane detection with self-supervised monocular depth estimation to predict explicit spatial information, without relying on expensive sensors or additional ground-truth depth data. 
    \item We show that our framework can be extended to learn camera intrinsics in a self-supervised manner. To address the instability of per-frame learned intrinsics, we propose a theoretically motivated fitting procedure to estimate parameters on a per-segment basis.
    \item We demonstrate that Depth3DLane outperforms previous methods, especially in terms of spatial accuracy. Furthermore, experimental results show that our method can be applied in scenarios where camera calibration is infeasible, unlike previous methods.
\end{enumerate}

\section{RELATED WORK}
\subsection{2D Lane Detection}
Monocular 2D lane detection methods predict lanes in the 2D image coordinate frame and can be categorized into four distinct approaches. \textit{Segmentation-based} methods pose lane detection as pixel-wise classification~\cite{EL-GAN2018, SAD2019, RESA2021}. However, lane masks are often incompatible with downstream tasks, so a separate curve-fitting stage is typically required~\cite{Neven2018, SCNN2018}. \textit{Curve-based} methods avoid intermediate lane masks and directly regress curve parameters to model lanes~\cite{Gansbeke2019, LSTR2021, Feng2022}. However, they suffer from slow convergence and typically fail to model complex lane geometry~\cite{GroupLane2023}. \textit{Keypoint-based} methods identify and cluster points of interest to predict lanes~\cite{Yoo2020, FOLOLane2021, GANet2022}. These methods can model complex lane geometries~\cite{RCLane2022}, but lack global structure priors, such as thinness and continuity, due to the semi-locality of keypoints~\cite{Qin2020}. \textit{Anchor-based} methods rely on techniques from generic anchor-based object detectors, such as YOLO~\cite{YOLO2016} and SSD~\cite{SSD2016}, to predict offsets for predefined anchor lanes~\cite{PointLaneNet2019, CondLaneNet2021, CLRNet2022, Laneformer2022}. These methods frequently achieve state-of-the-art performance, as they have more inductive bias than keypoint-based methods and are easier to optimize than curve-based methods~\cite{Line-CNN2019, LaneATT2021}.

Downstream tasks, such as lane departure warning, require lanes to be predicted in the 3D ego-vehicle space~\cite{Bar-HillelSurvey2014}. 2D methods rely on inverse perspective mapping (IPM)~\cite{IPM1991} to project lanes onto the 3D ground plane. However, IPM introduces non-negligible distortions when the flat ground assumption is broken, resulting in inaccurate 3D lane predictions for downstream tasks~\cite{Anchor3DLane2023, LATR2023}.

\subsection{3D Lane Detection}
Monocular 3D lane detection methods address the inherent limitations of 2D methods by predicting lanes directly in the 3D ego-vehicle coordinate frame. One of the first 3D methods, 3D-LaneNet~\cite{3D-LaneNet2019}, introduces a dual-pathway architecture that integrates IPM to project front view features to the bird's-eye view, where lanes are predicted using an anchor-based representation. Gen-LaneNet~\cite{Gen-LaneNet2020} further refines this approach by aligning anchors with visual features through a geometric transformation, improving performance in rarely observed scenes. PersFormer~\cite{PersFormer2022} further enhances robustness with a perspective transformer that leverages deformable attention~\cite{DeformableDETR2021} to refine correspondences between front view and bird's-eye view representations. To avoid the distortions caused by projecting to the bird's-eye view, Anchor3DLane~\cite{Anchor3DLane2023} proposes a lightweight CNN-based architecture using 3D lane anchors to sample features from the front view, achieving competitive performance at reduced computational cost. LATR~\cite{LATR2023} further improves upon this approach by adopting a transformer-based architecture~\cite{ViT2020} and introducing a dynamic ground plane to guide the learning of lane queries, resulting in state-of-the-art accuracy. Despite these advancements, monocular methods still struggle to accurately infer the 3D positions of lanes in complex scenarios, due to the inherent lack of explicit spatial information~\cite{Ma2024}.

To address the inherent lack of spatial information in monocular methods, multi-modal methods incorporate specialized depth sensors to provide explicit depth cues. For example, M$^2$-3DLane~\cite{M2-3DLaneNet2022} integrates LiDAR data to lift image features into the bird's-eye view without distortion, enhancing spatial accuracy. However, these depth sensors are prohibitively expensive for many applications, limiting their widespread adoption~\cite{Lite-Mono2023}. Alternatively, methods like SALAD~\cite{ONCE-3DLanes2022} rely on fully-supervised depth networks to provide explicit depth information, as this is typically more cost-effective. However, these methods rely on ground-truth depth data that is impractical to collect at scale~\cite{Zhou2017, Godard2019}.

Contrary to previous methods, Depth3DLane relies on self-supervised monocular depth estimation to predict explicit spatial information, without relying on expensive sensors or additional ground-truth depth data. Furthermore, the Depth3DLane framework can be extended to learn camera parameters in a self-supervised manner, enabling applications where camera calibration is infeasible, such as crowdsourced HD lane mapping~\cite{Crowdsourced3DTrafficSign2020}.

\begin{figure*}[t]
    \centering
    \vspace{6pt}
    \includegraphics[width=\textwidth]{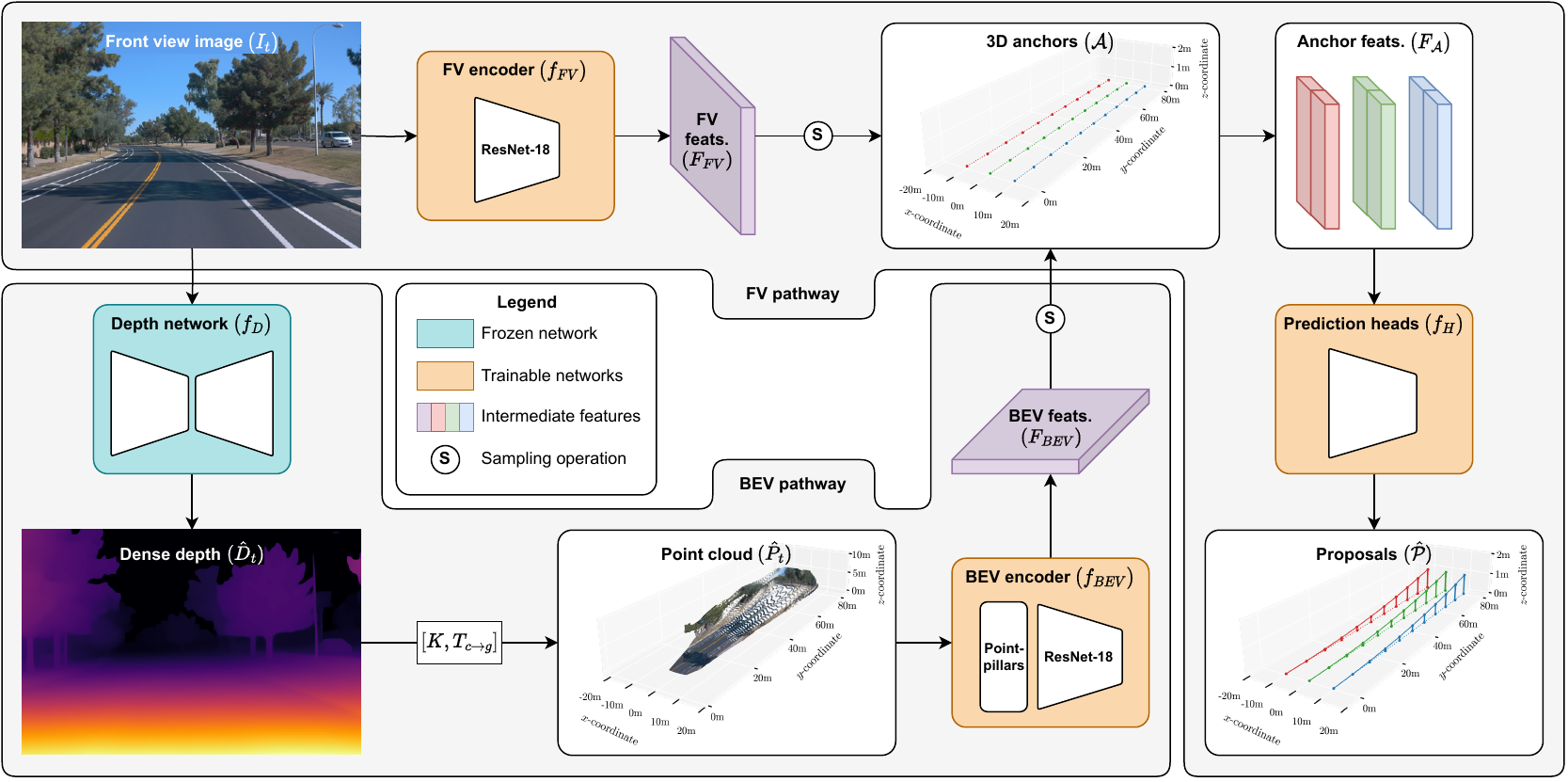}
    \caption{Graphical overview of our dual-pathway Depth3DLane framework.}
    \label{fig:depth3dlane}
\end{figure*}

\subsection{Self-Supervised Monocular Depth Estimation}
Specialized depth sensors provide explicit spatial information for 3D lane detection, but are prohibitively expensive for widespread adoption~\cite{Lite-Mono2023}. To avoid expensive sensors, deep learning-based learn to predict depth from monocular images~\cite{KITTIEigen2014}. However, fully-supervised methods require ground-truth depth data that is impractical to collect at scale, making self-supervised methods a promising alternative~\cite{Zhou2017}.

Self-supervised methods rely on two views of the scene to generate a supervisory signal. The core idea is that a target view can be reconstructed from a source view through view synthesis~\cite{Zhou2017}. To this end, a depth network predicts a dense depth map and a pose network predicts the relative camera pose between the two views, i.e., the ego-motion. By minimizing the photometric error between the original and reconstructed target views, the depth network learns to predict dense depth without relying on ground-truth depth data. After training, the depth and pose networks can be used independently of each other. 

In general, self-supervised methods assume that accurate camera parameters are available to perform view synthesis. However, this assumption precludes applications where camera calibration is infeasible, such as crowdsourced HD lane mapping~\cite{AIDrivenMaintenance2021}. To address this limitation, Gordon et al.~\cite{Gordon2019} and MT-SfMLearner~\cite{Varma2022} extend the pose network to predict the camera intrinsic matrix. This approach does not require changes to the optimization problem and enables applications where ground-truth camera parameters are not available.

Furthermore, monocular methods inherently suffer from scale-ambiguity, as infinitely many valid scenes can be obtained by simultaneously scaling object size and distance from the camera~\cite{Szeliski2022}. To address this issue, GPS-to-scale~\cite{Chawla2021} proposes an additional loss term that uses synchronized GPS data to constrain the scale of the predicted ego-motion, leveraging its pervasiveness in automotive applications. Importantly, this signal is not required during inference, as it only serves to constrain the predicted ego-motion during training, resulting in scale-aware features~\cite{Chawla2021}.

\section{METHOD}
\subsection{Overview}
Fig. \ref{fig:depth3dlane} provides a graphical overview of our dual-pathway Depth3DLane framework. Given input image $\mat{I}_t \in \R^{H \times W \times 3}$, the front view (FV) pathway extracts semantically rich features using a CNN-based encoder network $f_{FV}$. In parallel, the bird's-eye view (BEV) pathway leverages a self-supervised depth network $f_D$ to back-project the input image into a point cloud representation $\hat{\mat{P}}_t \in \R^{HW \times 8}$. Subsequently, a point cloud-based encoder network $f_{BEV}$ is utilized to extract spatial features from this representation. Following this, a set of 3D lane anchors is used to sample relevant features from $\mat{F}_{FV}$ and $\mat{F}_{BEV}$, integrating both spatial and semantic information. Finally, prediction heads $f_H$ produce the final set of lane proposals using an anchor-based representation.

\subsection{Scene Geometry and Lane Representation}
\begin{figure}[h]
    \centering
    \includegraphics[width=\columnwidth]{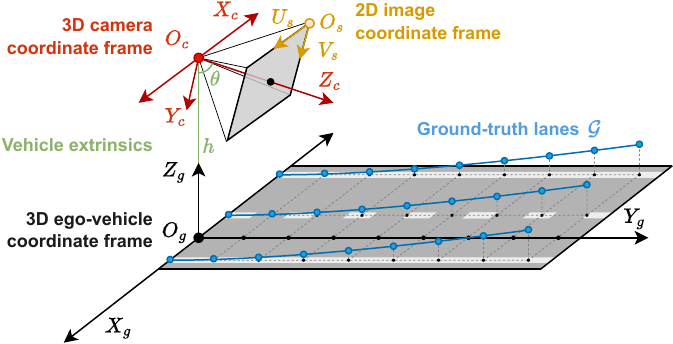}
    \caption{Schematic overview of the three coordinate systems relevant to monocular 3D lane detection.}
    \label{fig:geometry}
\end{figure}

Fig. \ref{fig:geometry} provides a graphical overview of the three coordinate frames that are relevant for monocular depth estimation and 3D lane detection. The 3D camera coordinate frame $(X_c, Y_c, Z_c)$ is defined at the camera center $O_c$, with $Y_c$ extending downward and~$Z_c$ extending perpendicular to the camera plane. The 3D ego-vehicle coordinate frame $(X_g, Y_g, Z_g)$ is defined at the projection $O_g$ of the camera center onto the ground plane, with $Z_g$ extending upward at the camera center and $Y_g$ extending forward in the driving direction. The 2D image coordinate frame $(U_s, V_s)$ is defined at the top-left corner~$O_s$ of the image sensor plane, with $U_s$ extending rightward and $V_s$ extending downward. 

Following previous methods~\cite{PersFormer2022, Anchor3DLane2023, LATR2023}, we represent lanes in the 3D ego-vehicle coordinate frame $O_g$ using a set of $N$ lane points at predefined $y$-coordinates $\vec{y} \in \R^N$. The set of ground-truth lanes $\set{G}$ contains $M_g$ lanes $G_i = (\vec{c}_i, \vec{x}_i, \vec{y}, \vec{z}_i, \vec{v}_i)$, where $\smash{\vec{c}_i \in \{0, 1\}^L}$ denotes the one-hot encoded lane category, $\smash{\vec{x}_i \in \R^N}$ and $\smash{\vec{z}_i \in \R^N}$ denote the $x$ and $z$-coordinates, respectively, and $\smash{\vec{v}_i \in [0, 1]^N}$ denotes a visibility flag to represent short or occluded lanes.

\subsection{Front View and Bird's-Eye View Pathways}
The front view pathway uses a CNN-based encoder network $\smash{f_{FV}}$ to extract semantically rich front view features $\smash{\mat{F}_{FV} \in \R^{H_{FV} \times W_{FV} \times d_{FV}}}$ from $\smash{\mat{I}_t}$. Simultaneously, the bird's-eye view pathway uses a self-supervised depth network $f_D$ to generate a point cloud representation $\smash{\hat{\mat{P}}_t \in \R^{HW \times 8}}$ from $\smash{\mat{I}_t}$. This point cloud is constructed by back-projecting every pixel into the 3D ego-vehicle coordinate frame and augmenting it with the corresponding color information and $uv$-coordinates for additional context. Subsequently, the bird's-eye view pathway uses a point cloud-based encoder network $f_{BEV}$ to extract bird's-eye view features $\smash{\mat{F}_{BEV} \in \R^{H_{BEV} \times W_{BEV} \times d_{BEV}}}$ from $\smash{\hat{\mat{P}}_t}$. To prevent $f_D$ from degrading during training, we pre-train the self-supervised depth model on a large set of unlabeled frames and GPS data, and freeze its weights when training the rest of the Depth3DLane framework.

\begin{table*}[t]
    \centering
    \vspace{6pt}
    \def\arraystretch{1.1}
    \caption{Comparison with state-of-the-art methods on the OpenLane validation datasets. \textbf{Bold} and \underline{underlined} text represent the best and second-best results respectively. The column \textbf{\#P} indicates the number of trainable parameters. The superscript \textdagger\ indicates that LATR uses input resolution 720 $\times$ 960 on OpenLane-1000, whereas Depth3DLane uses 320 $\times$ 480.}
    \begin{tabular}{lcccccccc}
        \thickhline
         \multirow{2}{*}{\textbf{Method}} & \multirow{2}{*}{\textbf{\#P} ($\downarrow$)} & \multicolumn{2}{c}{\textbf{Classification metrics} ($\uparrow$)} &  & \multicolumn{4}{c}{\textbf{Error metrics} ($\downarrow$)} \\ \cline{3-4} \cline{6-9} 
         &  & F1 score & Category acc. &  & X near (m) & X far (m) & Z near (m) & Z far (m) \\ \thickhline
        \multicolumn{9}{c}{OpenLane-1000 validation} \\ \hline
        3D-LaneNet~\cite{3D-LaneNet2019} & - & 44.1 & - &  & 0.479 & 0.572 & 0.367 & 0.443 \\
        Gen-LaneNet~\cite{Gen-LaneNet2020} & - & 32.3 & - &  & 0.591 & 0.684 & 0.411 & 0.521 \\
        PersFormer~\cite{PersFormer2022} & 54.9M & 50.5 & \textbf{92.3} &  & 0.485 & 0.553 & 0.364 & 0.431 \\
        CurveFormer~\cite{CurveFormer2022} & - & 50.5 & 89.5 &  & 0.319 & 0.325 & 0.112 & 0.141 \\
        CurveFormer++~\cite{CurveFormer++2024} & 27.4M & 52.5 & 87.8 &  & 0.333 & 0.805 & 0.186 & 0.687 \\
        Anchor3DLane~\cite{Anchor3DLane2023} & \textbf{13.3M} & 54.3 & 90.7 &  & 0.275 & 0.310 & 0.105 & 0.135 \\
        LATR\textsuperscript\textdagger~\cite{LATR2023} & 44.4M & \textbf{61.9} & \underline{92.0} &  & \textbf{0.219} & \textbf{0.259} & \underline{0.073} & \underline{0.106} \\
        \rowcolor[HTML]{e5e5e5} 
        Depth3DLane (ours) & \underline{26.9M} & \underline{56.6} & 86.1 &  & \underline{0.262} & \underline{0.300} & \textbf{0.068} & \textbf{0.105} \\ \thickhline
        \multicolumn{9}{c}{OpenLane-300 validation} \\ \hline
        LATR~\cite{LATR2023} & \underline{44.4M} & \underline{63.8} & \textbf{90.4} &  & \underline{0.310} & \underline{0.384} & \underline{0.113} & \underline{0.161} \\
        \rowcolor[HTML]{e5e5e5} 
        Depth3DLane (ours) & \textbf{26.9M} & \textbf{64.3} & \underline{87.0} &  & \textbf{0.289} & \textbf{0.371} & \textbf{0.091} & \textbf{0.131} \\ \thickhline
    \end{tabular}
    \label{tab:quantitative-results-openlane}
\end{table*}

\subsection{Self-Supervised Monocular Depth Estimation}
To train our self-supervised depth network, we adopt the general framework proposed by Zhou et al.~\cite{Zhou2017}. This framework relies on the task of novel view synthesis to jointly learn depth and ego-motion from a set of unlabeled video frames $\smash{\set{D} = \{ \mat{I}_1, ..., \mat{I}_T \}}$, where $\mat{I} \in \R^{H \times W \times 3}$ denotes a single frame. View synthesis reconstructs a target view $\mat{I}_t$ from a source view $\mat{I}_s$ using the following relationship between homogeneous image coordinates $\smash{\tilde{\vec{p}}_s \in \R^3}$ and $\smash{\tilde{\vec{p}}_t \in \R^3}$ in the source and target views, respectively:\footnote{The intermediate conversions between 3D and 4D homogeneous coordinates are left implicit for brevity.}
\begin{equation} \label{eq:view-synthesis}
    \tilde{\vec{p}}_s \sim \mat{K} \hat{\mat{T}}_{t \rightarrow s} \hat{\mat{D}}_t(\tilde{\vec{p}}_t) \mat{K}^{-1} \tilde{\vec{p}}_t,
\end{equation}
where $\hat{\mat{D}}_t \in \R^{H \times W}$ denotes the predicted pixel-wise depth from depth network $f_D$ and $\hat{\mat{T}}_{t \rightarrow s} \in \R^{4 \times 4}$ denotes the predicted ego-motion from pose network $f_D$. To obtain the reconstructed target view $\smash{\hat{\mat{I}}_{s \rightarrow t}}$, we sample continuous image coordinates $\vec{p}_s$ from the source view $\mat{I}_s$ through bilinear interpolation. 

Following Monodepth2~\cite{Godard2019} and Lite-Mono~\cite{Lite-Mono2023}, we jointly supervise the depth and pose networks using a combination of the photometric error and a smoothness loss. Additionally, we incorporate the GPS-to-scale loss~\cite{Chawla2021} to learn scale-aware features and predict depth at metric scale.

\subsection{3D Lane Anchors and Feature Sampling}
Inspired by Anchor3DLane~\cite{Anchor3DLane2023}, we use 3D lane anchors to sample features from the front view and bird's-eye view pathways. The set of anchors $\set{A}$ contains $M_a$ anchors $A_j = (\vec{x}_j, \vec{y}, \vec{z}_j)$, where $\vec{y} \in \R^N$ denotes the same predefined $y$-coordinates as for ground-truth lanes. To sample features from the front view, we project homogeneous 3D anchor points onto homogeneous 2D image coordinates as follows:
\begin{equation}
    \begin{bmatrix}
        \tilde{u}_s & \tilde{v}_s & \tilde{d}_s
    \end{bmatrix}^\top = \mat{K} \mat{T}_{g \rightarrow c} \begin{bmatrix}
        x_a & y_a & z_a & 1
    \end{bmatrix}^\top,
\end{equation}
and convert the resulting homogeneous image coordinates $(\tilde{u}_s, \tilde{v}_s, \tilde{d}_s)$ into normalized pixel coordinates $(\bar{u}_s, \bar{v}_s)$:
\begin{equation} \label{eq:homogeneous-conversion}
    \bar{u}_s = \frac{\tilde{u}_s}{W\tilde{d}_s}\ \text{ and }\ \bar{v}_s = \frac{\tilde{v}_s}{H\tilde{d}_s}.
\end{equation}

To sample features from the bird's-eye view, we use the predefined depth range $[y_{min}, y_{max}]$ and lateral extent $[x_{min}, x_{max}]$ to project 3D anchor points onto normalized ground plane coordinates $(\bar{x}_g, \bar{y}_g)$ as follows:
\begin{equation} \label{eq:homogeneous-conversion}
    \bar{x}_g = \frac{x_a - x_{min}}{x_{max} - x_{min}}\ \text{ and }\ \bar{y}_g = \frac{y_a - y_{min}}{y_{max} - y_{min}}.
\end{equation}

Finally, we use bilinear interpolation to sample front view features $F_{FV}(\bar{u}_s, \bar{v}_s)$ and bird's-eye view features $F_{BEV}(\bar{x}_g, \bar{y}_g)$. We then concatenate these features to obtain anchor features $\smash{\mat{F}_{\set{A}} \in \R^{M_a \times N \times (d_{FV} + d_{BEV})}}$.

\subsection{Prediction Heads and Losses}
We apply four prediction heads to the sampled anchor features $\mat{F}_\set{A}$ to produce the final set of proposals $\hat{\set{P}}$. The $x/z$-offset heads predict horizontal offsets $\smash{\Delta \hat{\vec{x}}_j \in \R^N}$ and vertical offsets $\smash{\Delta \hat{\vec{z}}_j \in \R^N}$, respectively, to adapt the position of the anchor in 3D space. Furthermore, the visibility head predicts a visibility vector $\smash{\hat{\vec{v}}_j \in \R^N}$ to represent short or partially occluded lanes. Finally, the classification head predicts a probability distribution $\hat{\vec{c}}_j \in [0, 1]^L$ over the $L$ lane categories. Given anchor $\smash{A_j = (\vec{x}_j, \vec{y}, \vec{z}_j)}$, the corresponding proposal $\smash{\hat{P}_j}$ is defined as follows:
\begin{equation} \label{eq:proposal}
    \hat{P}_j = (\hat{\vec{c}}_j,\ \vec{x}_j + \Delta \hat{\vec{x}}_j,\ \vec{y},\ \vec{z}_j + \Delta \hat{\vec{z}}_j,\ \hat{\vec{v}}_j).
\end{equation}

Following Anchor3DLane~\cite{Anchor3DLane2023}, we match $M_{pos}$ positive and $M_{neg}$ negative anchors to every ground-truth lane during training. We assign anchors instead of proposals to improve convergence during early stages of training, where proposals are still relatively noisy. The cost of matching anchor $A_j$ to ground-truth lane $G_i$ is defined as follows:
\begin{equation} \label{eq:matching-cost}
    D(G_i, A_j) = \frac{\sum_{k=1}^N v_i^k \sqrt{\left( x_i^k - x_j^k \right)^2 + \left( z_i^k - z_j^k \right)^2}}{\sum_{k=1}^N v_i^k}.
\end{equation}

To train the $x/z$-offset and visibility heads, we adopt the regression loss from previous methods~\cite{PersFormer2022, Anchor3DLane2023, LATR2023}. For every ground-truth lane $G_i$, we calculate the point-wise $L_1$ distance with the set of positive proposals $\smash{\hat{\set{P}}_{pos}(i)}$, which corresponds to the set of positive anchors $\set{A}_{pos}(i)$. For the $x/z$-offset losses, we divide the total point-wise distance by the number of visible points in the ground-truth lane. For the visibility loss, we average the total loss over all points. To train the classification head, we adopt the focal loss~\cite{FocalLoss2017}. For every ground-truth lane $G_i$, we pull the set of positive proposals $\smash{\hat{\set{P}}_{pos}(i)}$ towards the corresponding ground-truth class $\vec{c}_i$ and the set of negative proposals $\smash{\hat{\set{P}}_{neg}(i)}$ towards the background class~$\vec{0}$.

\subsection{Intrinsics Fitting Procedure}
Per-frame learned intrinsics are typically unstable, as the self-supervised optimization problem is highly underdetermined~\cite{Gordon2019}. To address this issue, we propose to fit intrinsics on a per-segment basis, as camera internals remain fixed between adjacent frames. To this end, we leverage the theoretical result from Gordon et al.~\cite{Gordon2019}, which states that per-frame learned focal lengths $\smash{\hat{f}_x, \hat{f}_y}$ are bounded by a function of true focal lengths $\smash{f_x, f_y}$, image dimensions $W, H$, and intra-frame rotations $r_z, r_x \in \R$, respectively.

Our fitting procedure adjust the focal length so that the theoretical bounds best fit the observed pairs of learned intrinsics and rotations across a segment. As learned intrinsics can exceed theoretical bounds and tend to underestimate ground-truth values~\cite{Gordon2019}, we propose to minimize the total amount by which per-frame learned intrinsics exceed the theoretical bounds. Given predictions for a driving segment $\smash{\set{D}_s = \{(\hat{r}^1_z, \hat{f}^1_ x), ..., (\hat{r}^T_z, \hat{f}^T_ x)\}}$, we fit the per-segment focal length $\smash{\hat{f}^s_x \in \R^+}$ as follows, using the $ReLU$ function to mask points that do not exceed the theoretical bounds:
\begin{equation} \label{eq:lane-fitting-objective}
    \hat{f}^s_x = \min_{f_x} \sum_{i=1}^T ReLU\left( |f_x - \hat{f}^i_x| - \frac{2f^2_x}{W^2\hat{r}_z} \right).
\end{equation}

We assume that $\smash{\hat{f}^s_x = \hat{f}^s_y}$, as automotive datasets feature limited vertical rotation to infer $\smash{\hat{f}^s_y}$ from (see Sec. \ref{sec:ablation}).

\section{EXPERIMENTS}
\begin{table*}[t]
    \centering
    \vspace{6pt}
    \def\arraystretch{1.1}
    \caption{Comparison of different configurations of the bird's-eye view pathway on the OpenLane-300 validation dataset. \textbf{Bold} text represents the best results. PP and RN18 indicate the PointPillars and ResNet-18 networks, respectively.}
    \begin{tabular}{lccccccc}
        \thickhline
        \multirow{2}{*}{\textbf{BEV pathway}} & \multicolumn{2}{c}{\textbf{Classification metrics} ($\uparrow$)} &  & \multicolumn{4}{c}{\textbf{Error metrics} ($\downarrow$)} \\ \cline{2-3} \cline{5-8} 
         & F1 score & Category acc. &  & X near (m) & X far (m) & Z near (m) & Z far (m) \\ \thickhline
        No BEV pathway & 61.7 & 86.1 &  & 0.309 & 0.392 & 0.095 & 0.135 \\
        BEV pathway (PP) & 62.6 & 86.5 &  & 0.299 & 0.383 & 0.093 & \textbf{0.131} \\
        BEV pathway (PP + RN18) & \textbf{64.3} & \textbf{87.0} &  & \textbf{0.289} & \textbf{0.371} & \textbf{0.091} & \textbf{0.131} \\ \thickhline
    \end{tabular}
    \label{tab:ablation-bev-encoder}
\end{table*}

\begin{table*}[t]
    \centering
    \def\arraystretch{1.1}
    \caption{\footnotesize Comparison of different intrinsics learning methods on the OpenLane-300 validation dataset. \textbf{Bold} text represents the best results. $r_{z}^{min}$ denotes the minimum horizontal rotation used to fit the focal length $f$.}
    \begin{tabular}{lccccccccc}
        \thickhline
        \multirow{2}{*}{\textbf{Intrinsics}} & \multirow{2}{*}{$r_z^{min}$} & \multirow{2}{*}{$f$} & \multicolumn{2}{c}{\textbf{Classification metrics} ($\uparrow$)} &  & \multicolumn{4}{c}{\textbf{Error metrics} ($\downarrow$)} \\ \cline{4-5} \cline{7-10} 
         &  &  & F1 score & Category acc. &  & X near (m) & X far (m) & Z near (m) & Z far (m) \\ \thickhline
        Fixed & - & 516 & \textbf{64.3} & \textbf{87.0} &  & \textbf{0.289} & \textbf{0.371} & \textbf{0.091} & \textbf{0.131} \\
        Learned & 0.01 & 446 & 60.8 & 86.0 &  & 0.353 & 0.517 & 0.119 & 0.197 \\
        Learned & 0.02 & 506 & 60.4 & 86.2 &  & 0.348 & 0.503 & 0.113 & 0.189 \\
        Learned & 0.03 & 517 & 61.1 & 85.5 &  & 0.334 & 0.492 & 0.118 & 0.190 \\ \thickhline
    \end{tabular}
    \label{tab:ablation-camera-parameters}
\end{table*}

\subsection{Dataset and Evaluation Metrics}
\textbf{OpenLane}~\cite{PersFormer2022} is a large-scale, real-world 3D lane detection benchmark built on the Waymo Open dataset~\cite{WaymoOpen2020}. In total, the dataset comprises 880,000 lane annotations across 200,000 frames and 14 lane categories. Frames are captured under realistic driving conditions, including extreme weather and night driving. Camera intrinsics and extrinsics are available for every frame. Following Anchor3DLane~\cite{Anchor3DLane2023} and CurveFormer++~\cite{CurveFormer2022}, we extend the dataset to include relative camera poses to train our self-supervised depth network with the GPS-to-scale loss~\cite{Chawla2021}.

\textbf{Evaluation Metrics.} We follow the official evaluation metrics from Gen-LaneNet~\cite{Gen-LaneNet2020} to evaluate our method on the OpenLane dataset. First, a minimum-cost flow problem is solved to match proposals to ground-truth lanes. The cost of matching two lanes is defined as the square root of the sum of point-wise Euclidean distances, with a penalty of 1.5m for points with mispredicted visibility. Two lanes are considered matched if at least 75\% of the point-wise distances are smaller than 1.5m. We report the F1 score and category accuracy to evaluate lane detection accuracy, and $x/z$-errors close to ($<$40m) and far from ($>$40m) the camera to evaluate spatial accuracy.

\subsection{Implementation Details}
We implement our Depth3DLane model in PyTorch and perform experiments on a single NVIDIA RTX 3080Ti. We first train our self-supervised depth network on frames and poses from the OpenLane dataset using a batch size of~12. We use the AdamW optimizer~\cite{AdamW2017} with a learning rate of $1e^{-4}$ and weight decay of $1e^{-2}$, training for 35 epochs and halving the learning rate after 20 epochs. We adopt the depth network architecture from Lite-Mono~\cite{Lite-Mono2023} and the pose network architecture from MT-SfMLearner~\cite{Varma2022}. 

After training the depth network, we freeze its weights and train the rest of our Depth3DLane model with a batch size of 6. We use the Adam optimizer~\cite{Adam2015} with a learning rate of $2e^{-4}$ and train for 25 epochs using a cosine annealing schedule with a minimum learning rate of $2e^{-5}$, a decay rate of 0.95, and a cycle length of 3 epochs. We use a modified ResNet-18~\cite{ResNet2016} for the front view pathway and a PointPillars~\cite{PointPillars2019} model with an additional ResNet-18 for the bird's-eye view pathway. For experiments on the OpenLane dataset, we use an input resolution of $320 \times 480$, a minimum depth of 0.1m, a maximum depth of 80m, and a maximum lateral extent of 20m. We cast anchors from 45 uniformly spaced points along the $X_g$ axis, up to the maximum lateral extent, and at pitch angles $\{0^\circ, \pm 1^\circ, \pm 2 ^\circ\}$ and yaw angles $\{0^\circ, \pm1^\circ, \pm3^\circ, \pm5^\circ, \pm7^\circ, \pm10^\circ, \pm15^\circ, \pm20^\circ\}$.

\subsection{Quantitative Results}
Table \ref{tab:quantitative-results-openlane} presents results on the OpenLane-1000 and OpenLane-300 validation datasets for the base configuration of Depth3DLane. As shown, the overall performance on the OpenLane-1000 dataset is competitive with state-of-the-art methods. Notably, our model exhibits remarkable spatial accuracy, achieving the lowest $z$-error metrics and second-lowest $x$-error metrics. Additionally, Depth3DLane requires fewer parameters than PersFormer~\cite{PersFormer2022}, CurveFormer++~\cite{CurveFormer++2024}, and LATR~\cite{LATR2023}, further highlighting the effectiveness of integrating explicit spatial information. In terms of classification metrics, our model does not surpass the current state-of-the-art model LATR. However, the base configuration of LATR uses a significantly higher input resolution, which primarily affects classification performance~\cite{LATR2023}, preventing a fair direct comparison. At a similar resolution, our model outperforms LATR on the OpenLane-300 validation dataset. For a more direct comparison on the OpenLane-1000 dataset, we instead refer to Anchor3DLane~\cite{Anchor3DLane2023}, which uses a comparable input resolution and is similar to the front view pathway in Depth3DLane. In particular, we improve the F1 score of Anchor3DLane by 4.3\% (54.3~$\rightarrow$~56.6), highlighting the effectiveness of the bird's-eye view pathway.

\subsection{Ablation Studies}
\label{sec:ablation}
\textbf{BEV Pathway.} We perform an ablation study on the BEV pathway to verify the importance of adding explicit spatial information. Table~\ref{tab:ablation-bev-encoder} presents results on the OpenLane-300 validation dataset for three configurations of the BEV pathway. As shown, the base configuration with PointPillars and ResNet-18 improves the F1 score by 4.3\% (61.7~$\rightarrow$~64.3) and reduces error metrics by up to 6.5\% (0.309~$\rightarrow$~0.289). Moreover, the second configuration with only PointPillars to extract basic spatial features, also shows improved performance. From these results, we conclude that the bird's-eye view pathway effectively enhances performance by integrating explicit spatial information into the model.

\textbf{Learned Intrinsics.} We perform additional experiments to assess the effectiveness of our intrinsics fitting procedure. Table~\ref{tab:ablation-camera-parameters} presents results on the OpenLane-300 validation dataset for three configurations with learned intrinsics. As shown, we can improve our estimate of the ground-truth focal length by filtering out instances where the horizontal rotation $r_z$ is less than a predefined threshold $\smash{r_z^{min}}$. In particular, we approximate the mean ground-truth focal length of 516 using a threshold of 0.03 radians.

Naturally, configurations with learned intrinsics perform worse than the baseline model with ground-truth intrinsics. This is likely due to our assumption that $f_x = f_y$, as there is practically no vertical rotation in automotive datasets to accurately fit $f_y$ separately. This assumption does not always hold, leading to misaligned features when projecting anchors to the front view using inaccurate camera parameters. Nevertheless, this approach allows our method to be applied in scenarios where camera calibration is infeasible, unlike previous methods.

\section{CONCLUSION}
Our Depth3DLane framework effectively combines rich semantic information from the front view pathway with explicit spatial information from the bird's-eye view pathway to accurately infer 3D lane geometry. Experimental results show that Depth3DLane outperforms previous methods, especially in spatial accuracy. Furthermore, we show that the framework can be extended to learn camera intrinsics in a self-supervised manner. Our proposed fitting procedure addresses the instability of per-frame learned intrinsics and approximates ground-truth focal lengths on a per-segment basis. Experimental results show that this approach allows Depth3DLane to be applied in scenarios where camera calibration is infeasible.

\bibliographystyle{IEEEtran}  
\bibliography{references}

\begin{thebibliography}{10}
\providecommand{\url}[1]{#1}
\csname url@rmstyle\endcsname
\providecommand{\newblock}{\relax}
\providecommand{\bibinfo}[2]{#2}
\providecommand\BIBentrySTDinterwordspacing{\spaceskip=0pt\relax}
\providecommand\BIBentryALTinterwordstretchfactor{4}
\providecommand\BIBentryALTinterwordspacing{\spaceskip=\fontdimen2\font plus
\BIBentryALTinterwordstretchfactor\fontdimen3\font minus \fontdimen4\font\relax}
\providecommand\BIBforeignlanguage[2]{{%
\expandafter\ifx\csname l@#1\endcsname\relax
\typeout{** WARNING: IEEEtran.bst: No hyphenation pattern has been}%
\typeout{** loaded for the language `#1'. Using the pattern for}%
\typeout{** the default language instead.}%
\else
\language=\csname l@#1\endcsname
\fi
#2}}

\bibitem{Narote2018}
S.~P. Narote, P.~N. Bhujbal, A.~S. Narote, and D.~M. Dhane, ``{A review of recent advances in lane detection and departure warning system},'' \emph{Pattern Recognition}, vol.~73, pp. 216--234, 2018.

\bibitem{Liu2020}
R.~Liu, J.~Wang, and B.~Zhang, ``{High Definition Map for Automated Driving: Overview and Analysis},'' \emph{The Journal of Navigation}, vol.~73, pp. 324--341, 2020.

\bibitem{AIDrivenMaintenance2021}
R.~Mukherjee, H.~Iqbal, S.~Marzban, A.~Badar, T.~Brouns, S.~Gowda, E.~Arani, and B.~Zonooz, ``{AI-Driven Road Maintenance Inspection},'' \emph{ITS World Congress}, pp. 11--15, 2021.

\bibitem{AIDrivenMaintenanceV22022}
H.~Iqbal, H.~Chawla, A.~Varma, T.~Brouns, A.~Badar, E.~Arani, and B.~Zonooz, ``{AI-Driven Road Maintenance Inspection v2: Reducing Data Dependency \& Quantifying Road Damage},'' \emph{IRF Global R2T Conference \& Exhibition}, 2022.

\bibitem{BEV-LaneDet2023}
R.~Wang, J.~Qin, K.~Li, Y.~Li, D.~Cao, and J.~Xu, ``{BEV-LaneDet: An Efficient 3D Lane Detection Based on Virtual Camera via Key-Points},'' pp. 1002--1011, 2023.

\bibitem{IPM1991}
H.~A. Mallot, H.~H. Bülthoff, J.~J. Little, and S.~Bohrer, ``{Inverse perspective mapping simplifies optical flow computation and obstacle detection},'' \emph{Biological Cybernetics}, vol.~64, pp. 177--185, 1991.

\bibitem{3D-LaneNet2019}
N.~Garnett, R.~Cohen, T.~Pe'Er, R.~Lahav, and D.~Levi, ``{3D-LaneNet: End-to-End 3D Multiple Lane Detection},'' \emph{International Conference on Computer Vision}, pp. 2921--2930, 2019.

\bibitem{ReconstructFromBEV2022}
C.~Li, J.~Shi, Y.~Wang, and G.~Cheng, ``{Reconstruct from BEV: A 3D Lane Detection Approach based on Geometry Structure Prior},'' \emph{Conference on Computer Vision and Pattern Recognition}, pp. 4369--4378, 2022.

\bibitem{Anchor3DLane2023}
S.~Huang, Z.~Shen, Z.~Huang, Z.~han Ding, J.~Dai, J.~Han, N.~Wang, and S.~Liu, ``{Anchor3DLane: Learning to Regress 3D Anchors for Monocular 3D Lane Detection},'' \emph{Conference on Computer Vision and Pattern Recognition}, pp. 17\,451--17\,460, 2023.

\bibitem{LATR2023}
Y.~Luo, C.~Zheng, X.~Yan, T.~Kun, C.~Zheng, S.~Cui, and Z.~Li, ``{LATR: 3D Lane Detection from Monocular Images with Transformer},'' \emph{International Conference on Computer Vision}, pp. 7941--7952, 2023.

\bibitem{Gen-LaneNet2020}
Y.~Guo, G.~Chen, P.~Zhao, W.~Zhang, J.~Miao, J.~Wang, and T.~E. Choe, ``{Gen-LaneNet: A Generalized and Scalable Approach for 3D Lane Detection},'' \emph{European Conference on Computer Vision}, pp. 666--681, 2020.

\bibitem{M2-3DLaneNet2022}
Y.~Luo, X.~Yan, C.~Zheng, C.~Zheng, S.~Mei, T.~Kun, S.~Cui, Z.~Li, T.~Map, and T.~Lab, ``{M$^2$-3DLaneNet: Exploring Multi-Modal 3D Lane Detection},'' \emph{arXiv preprint arXiv:2209.05996v3}, 2022.

\bibitem{Lite-Mono2023}
N.~Zhang, F.~Nex, G.~Vosselman, and N.~Kerle, ``{Lite-Mono: A Lightweight CNN and Transformer Architecture for Self-Supervised Monocular Depth Estimation},'' \emph{Conference on Computer Vision and Pattern Recognition}, pp. 18\,537--18\,546, 2023.

\bibitem{ONCE-3DLanes2022}
F.~Yan, M.~Nie, X.~Cai, J.~Han, H.~Xu, Z.~Yang, C.~Ye, Y.~Fu, M.~B. Mi, and L.~Zhang, ``{ONCE-3DLanes: Building Monocular 3D Lane Detection},'' \emph{Conference on Computer Vision and Pattern Recognition}, pp. 17\,143--17\,152, 2022.

\bibitem{Zhou2017}
T.~Zhou, M.~Brown, N.~Snavely, and D.~G. Lowe, ``{Unsupervised Learning of Depth and Ego-Motion from Video},'' \emph{Conference on Computer Vision and Pattern Recognition}, pp. 6612--6621, 2017.

\bibitem{Godard2019}
C.~Godard, O.~M. Aodha, M.~Firman, and G.~Brostow, ``{Digging Into Self-Supervised Monocular Depth Estimation},'' \emph{International Conference on Computer Vision}, pp. 3827--3837, 2019.

\bibitem{Ma2024}
F.~Ma, W.~Qi, G.~Zhao, L.~Zheng, S.~Wang, Y.~Liu, and M.~Liu, ``{Monocular 3D lane detection for Autonomous Driving: Recent Achievements, Challenges, and Outlooks},'' \emph{arXiv preprint arXiv:2404.06860v2}, 2024.

\bibitem{EL-GAN2018}
M.~Ghafoorian, C.~Nugteren, N.~Baka, O.~Booij, and M.~Hofmann, ``{EL-GAN: Embedding Loss Driven Generative Adversarial Networks for Lane Detection},'' \emph{European Conference on Computer Vision}, pp. 256--272, 2018.

\bibitem{SAD2019}
Y.~Hou, Z.~Ma, C.~Liu, and C.~C. Loy, ``{Learning Lightweight Lane Detection CNNs by Self Attention Distillation},'' \emph{International Conference on Computer Vision}, pp. 1013--1021, 2019.

\bibitem{RESA2021}
T.~Zheng, H.~Fang, Y.~Zhang, W.~Tang, Z.~Yang, H.~F. Liu, and D.~Cai, ``{RESA: Recurrent Feature-Shift Aggregator for Lane Detection},'' \emph{Conference on Artificial Intelligence}, vol.~4B, pp. 3547--3554, 2021.

\bibitem{Neven2018}
D.~Neven, B.~D. Brabandere, S.~Georgoulis, M.~Proesmans, and L.~V. Gool, ``{Towards End-to-End Lane Detection: an Instance Segmentation Approach},'' \emph{Intelligent Vehicles Symposium}, pp. 286--291, 2018.

\bibitem{SCNN2018}
X.~Pan, J.~Shi, P.~Luo, X.~Wang, and X.~Tang, ``{Spatial As Deep: Spatial CNN for Traffic Scene Understanding},'' \emph{Conference on Artificial Intelligence}, pp. 7276--7283, 2018.

\bibitem{Gansbeke2019}
W.~V. Gansbeke, B.~D. Brabandere, D.~Neven, M.~Proesmans, and L.~V. Gool, ``{End-to-end Lane Detection through Differentiable Least-Squares Fitting},'' \emph{International Conference on Computer Vision Workshop}, pp. 905--913, 2019.

\bibitem{LSTR2021}
R.~Liu, Z.~Yuan, T.~Liu, and Z.~Xiong, ``{End-to-end Lane Shape Prediction with Transformers},'' \emph{Winter Conference on Applications of Computer Vision}, pp. 3693--3701, 2021.

\bibitem{Feng2022}
Z.~Feng, S.~Guo, X.~Tan, K.~Xu, M.~Wang, and L.~Ma, ``{Rethinking Efficient Lane Detection via Curve Modeling},'' \emph{Conference on Computer Vision and Pattern Recognition}, pp. 17\,041--17\,049, 2022.

\bibitem{GroupLane2023}
Z.~Li, C.~Han, Z.~Ge, J.~Yang, E.~Yu, H.~Wang, H.~Zhao, and X.~Zhang, ``{GroupLane: End-to-End 3D Lane Detection with Channel-wise Grouping},'' \emph{arXiv preprint arXiv:2307.09472v1}, 2023.

\bibitem{Yoo2020}
S.~Yoo, H.~S. Lee, H.~Myeong, S.~Yun, H.~Park, J.~Cho, and D.~H. Kim, ``{End-to-End Lane Marker Detection via Row-wise Classification},'' \emph{Conference on Computer Vision and Pattern Recognition}, pp. 4335--4343, 2020.

\bibitem{FOLOLane2021}
Z.~Qu, H.~Jin, Y.~Zhou, Z.~Yang, and W.~Zhang, ``{Focus on Local: Detecting Lane Marker from Bottom Up via Key Point},'' \emph{Conference on Computer Vision and Pattern Recognition}, pp. 14\,117--14\,125, 2021.

\bibitem{GANet2022}
J.~Wang, Y.~Ma, S.~Huang, T.~Hui, F.~Wang, C.~Qian, and T.~Zhang, ``{A Keypoint-based Global Association Network for Lane Detection},'' \emph{Conference on Computer Vision and Pattern Recognition}, pp. 1382--1391, 2022.

\bibitem{RCLane2022}
S.~Xu, X.~Cai, B.~Zhao, L.~Zhang, H.~Xu, Y.~Fu, and X.~Xue, ``{RCLane: Relay Chain Prediction for Lane Detection},'' \emph{European Conference on Computer Vision}, vol. 13698, pp. 461--477, 2022.

\bibitem{Qin2020}
Z.~Qin, H.~Wang, and X.~Li, ``{Ultra Fast Structure-aware Deep Lane Detection},'' \emph{European Conference on Computer Vision}, vol. 12369, pp. 276--291, 2020.

\bibitem{YOLO2016}
J.~Redmon, S.~Divvala, R.~Girshick, and A.~Farhadi, ``{You Only Look Once: Unified, Real-Time Object Detection},'' \emph{Conference on Computer Vision and Pattern Recognition}, pp. 779--788, 2016.

\bibitem{SSD2016}
W.~Liu, D.~Anguelov, D.~Erhan, C.~Szegedy, S.~Reed, C.-Y. Fu, and A.~C. Berg, ``{SSD: Single Shot MultiBox Detector},'' \emph{European Conference on Computer Vision}, vol. 9905, pp. 21--37, 2016.

\bibitem{PointLaneNet2019}
Z.~Chen, Q.~Liu, and C.~Lian, ``{PointLaneNet: Efficient end-to-end CNNs for accurate real-time lane detection},'' \emph{Intelligent Vehicles Symposium}, pp. 2563--2568, 2019.

\bibitem{CondLaneNet2021}
L.~Liu, X.~Chen, S.~Zhu, and P.~Tan, ``{CondLaneNet: a Top-to-down Lane Detection Framework Based on Conditional Convolution},'' \emph{International Conference on Computer Vision}, pp. 3753--3762, 2021.

\bibitem{CLRNet2022}
T.~Zheng, Y.~Huang, Y.~Liu, W.~Tang, Z.~Yang, D.~Cai, and X.~He, ``{CLRNet: Cross Layer Refinement Network for Lane Detection},'' \emph{Conference on Computer Vision and Pattern Recognition}, pp. 888--897, 2022.

\bibitem{Laneformer2022}
J.~Han, X.~Deng, X.~Cai, Z.~Yang, H.~Xu, C.~Xu, and X.~Liang, ``{Laneformer: Object-aware Row-Column Transformers for Lane Detection},'' \emph{Conference on Artificial Intelligence}, vol.~36, pp. 1122--1130, 2022.

\bibitem{Line-CNN2019}
X.~Li, J.~Li, X.~Hu, and J.~Yang, ``{Line-CNN: End-to-End Traffic Line Detection with Line Proposal Unit},'' \emph{Intelligent Transportation Systems}, vol.~21, pp. 248--258, 2019.

\bibitem{LaneATT2021}
L.~Tabelini, R.~Berriel, T.~M. Paixão, C.~Badue, A.~F. de~Souza, and T.~Oliveira-Santos, ``{Keep your Eyes on the Lane: Real-time Attention-guided Lane Detection},'' \emph{Conference on Computer Vision and Pattern Recognition}, pp. 294--302, 2021.

\bibitem{Bar-HillelSurvey2014}
A.~B. Hillel, R.~Lerner, D.~Levi, and G.~Raz, ``{Recent progress in road and lane detection: A survey},'' \emph{Machine Vision and Applications}, vol.~25, pp. 727--745, 2014.

\bibitem{PersFormer2022}
L.~Chen, C.~Sima, Y.~Li, Z.~Zheng, J.~Xu, X.~Geng, H.~Li, C.~He, J.~Shi, Y.~Qiao, and J.~Yan, ``{PersFormer: 3D Lane Detection via Perspective Transformer and the OpenLane Benchmark},'' \emph{European Conference on Computer Vision}, pp. 550--567, 2022.

\bibitem{DeformableDETR2021}
X.~Zhu, W.~Su, L.~Lu, B.~Li, X.~Wang, and J.~Dai, ``{Deformable DETR: Deformable Transformers for End-to-End Object Detection},'' \emph{International Conference on Learning Representations}, 2021.

\bibitem{ViT2020}
A.~Dosovitskiy, L.~Beyer, A.~Kolesnikov, D.~Weissenborn, X.~Zhai, T.~Unterthiner, M.~Dehghani, M.~Minderer, G.~Heigold, S.~Gelly, J.~Uszkoreit, and N.~Houlsby, ``{An Image is Worth 16x16 Words: Transformers for Image Recognition at Scale},'' \emph{International Conference on Learning Representations}, 2020.

\bibitem{Crowdsourced3DTrafficSign2020}
H.~Chawla, M.~Jukola, E.~Arani, and B.~Zonooz, ``{Monocular Vision based Crowdsourced 3D Traffic Sign Positioning with Unknown Camera Intrinsics and Distortion Coefficients},'' \emph{International Conference on Intelligent Transportation Systems}, 2020.

\bibitem{KITTIEigen2014}
D.~Eigen, C.~Puhrsch, and R.~Fergus, ``{Depth Map Prediction from a Single Image using a Multi-Scale Deep Network},'' \emph{Advances in Neural Information Processing Systems}, vol.~3, pp. 2366--2374, 2014.

\bibitem{Gordon2019}
A.~Gordon, H.~Li, R.~Jonschkowski, and A.~Angelova, ``{Depth from Videos in the Wild: Unsupervised Monocular Depth Learning from Unknown Cameras},'' \emph{International Conference on Computer Vision}, pp. 8976--8985, 2019.

\bibitem{Varma2022}
A.~Varma, H.~Chawla, B.~Zonooz, and E.~Arani, ``{Transformers in Self-Supervised Monocular Depth Estimation with Unknown Camera Intrinsics},'' \emph{International Conference on Computer Vision Theory and Applications}, pp. 758--769, 2022.

\bibitem{Szeliski2022}
R.~Szeliski, \emph{{Computer Vision}}, 2022.

\bibitem{Chawla2021}
H.~Chawla, A.~Varma, E.~Arani, and B.~Zonooz, ``{Multimodal Scale Consistency and Awareness for Monocular Self-Supervised Depth Estimation},'' \emph{International Conference on Robotics and Automation}, pp. 5140--5146, 2021.

\bibitem{CurveFormer2022}
Y.~Bai, Z.~Chen, Z.~Fu, L.~Peng, P.~Liang, and E.~Cheng, ``{CurveFormer: 3D Lane Detection by Curve Propagation with Curve Queries and Attention},'' \emph{International Conference on Robotics and Automation}, pp. 7062--7068, 2022.

\bibitem{CurveFormer++2024}
Y.~Bai, Z.~Chen, P.~Liang, and E.~Cheng, ``{CurveFormer++: 3D Lane Detection by Curve Propagation with Temporal Curve Queries and Attention},'' \emph{arXiv preprint arXiv:2402.06423v1}, 2024.

\bibitem{FocalLoss2017}
T.~Y. Lin, P.~Goyal, R.~Girshick, K.~He, and P.~Dollar, ``{Focal Loss for Dense Object Detection},'' \emph{Pattern Analysis and Machine Intelligence}, vol.~42, pp. 318--327, 2017.

\bibitem{WaymoOpen2020}
P.~Sun, H.~Kretzschmar, X.~Dotiwalla, A.~Chouard, V.~Patnaik, P.~Tsui, J.~Guo, Y.~Zhou, Y.~Chai, B.~Caine, V.~Vasudevan, W.~Han, J.~Ngiam, H.~Zhao, A.~Timofeev, S.~Ettinger, M.~Krivokon, A.~Gao, A.~Joshi, Y.~Zhang, J.~Shlens, Z.~Chen, and D.~Anguelov, ``{Scalability in Perception for Autonomous Driving: Waymo Open Dataset},'' \emph{Conference on Computer Vision and Pattern Recognition}, pp. 2443--2451, 2020.

\bibitem{AdamW2017}
I.~Loshchilov and F.~Hutter, ``{Decoupled Weight Decay Regularization},'' \emph{International Conference on Learning Representations}, 2017.

\bibitem{Adam2015}
D.~P. Kingma and J.~L. Ba, ``{Adam: A Method for Stochastic Optimization},'' \emph{International Conference on Learning Representations}, 2015.

\bibitem{ResNet2016}
K.~He, X.~Zhang, S.~Ren, and J.~Sun, ``{Deep Residual Learning for Image Recognition},'' \emph{Conference on Computer Vision and Pattern Recognition}, pp. 770--778, 2016.

\bibitem{PointPillars2019}
A.~H. Lang, S.~Vora, H.~Caesar, L.~Zhou, J.~Yang, and O.~Beijbom, ``{PointPillars: Fast Encoders for Object Detection from Point Clouds},'' \emph{Conference on Computer Vision and Pattern Recognition}, pp. 12\,689--12\,697, 2019.

\end{thebibliography}

\end{document}